\documentclass[lettersize,journal]{IEEEtran}
\usepackage{amsmath,amsfonts}
\usepackage[noend]{algorithmic}
\usepackage{algorithm}
\usepackage{array}
\usepackage[caption=false,font=normalsize,labelfont=sf,textfont=sf]{subfig}
\usepackage{textcomp}
\usepackage{stfloats}
\usepackage{url}
\usepackage{verbatim}
\usepackage{graphicx}
\usepackage{subfig} 
\usepackage{cite}
\usepackage{multirow}
\usepackage{xcolor}  
\usepackage{array} 
\usepackage{enumitem} 
\usepackage{colortbl} 

\begin{document}

\title{Advanced YOLO-based Real-time Power Line Detection for Vegetation Management}

\author{Shuaiang Rong,~\IEEEmembership{Student Member,~IEEE}, Lina He, ~\IEEEmembership{Senior Member,~IEEE}, Salih Furkan Atici, Ahmet Enis Cetin,~\IEEEmembership{Fellow,~IEEE} 
 
\thanks{This work is funded in part by USDA Forest Service 11261925.}
        
\thanks{Shuaiang Rong, Lina He, and Ahmet Enis Cetin are with the Electrical and Computer Engineering Department, University of Illinois at Chicago, Chicago, IL 60607, USA (e-mail: srong4@uic.edu; lhe@uic.edu; aecyy@uic.edu).}
\thanks{Salih Furkan Atici is with Apple Inc., Cupertino, CA 95014, USA (e-mail: satici2@uic.edu).}
\thanks{Manuscript received XXX; revised 30 January 2025.}}

\markboth{Journal of \LaTeX\ Class Files,~Vol.XX, No.XX, XX~2024}%
{Shell \MakeLowercase{\textit{et al.}}: A Sample Article Using IEEEtran.cls for IEEE Journals}

\IEEEpubid{0000--0000/00\$00.00~\copyright~2021 IEEE}

\maketitle

\begin{abstract}
Power line infrastructure is a key component of the power system, and it is rapidly expanding to meet growing energy demands. Vegetation encroachment is a significant threat to the safe operation of power lines, requiring reliable and timely management to enhance the resilience and reliability of the power network. Integrating smart grid technology, especially Unmanned Aerial Vehicles (UAVs), provides substantial potential to revolutionize the management of extensive power line networks with advanced imaging techniques. However, processing the vast quantity of images captured by UAV patrols remains a significant challenge. This paper introduces an intelligent real-time monitoring framework for detecting power lines and adjacent vegetation. It is developed based on the deep-learning Convolutional Neural Network (CNN), You Only Look Once (YOLO), renowned for its high-speed object detection capabilities. Unlike existing deep learning-based methods, this framework enhances accuracy by integrating YOLOv8 with directional filters. They can extract directional features and textures of power lines and their vicinity, generating Oriented Bounding Boxes (OBB) for more precise localization. Additionally, a post-processing algorithm is developed to create a vegetation encroachment metric for power lines, allowing for a quantitative assessment of the surrounding vegetation distribution. The effectiveness of the proposed framework is demonstrated using a widely used power line dataset. 
\end{abstract}

\begin{IEEEkeywords}
Power line detection, monitoring, CNN, YOLOv8, vegetation management, UAV.
\end{IEEEkeywords}

\section{Introduction}
\IEEEPARstart{P}{ower} lines can produce sparks upon tree contact due to factors such as line sag, tree growth, and severe weather conditions during natural hazards. Tree-contacted power line sparking has been identified as a significant cause of the most destructive wildfires in California in recent years \cite{ref1}, as well as the devastating Maui fire in 2023 \cite{ref2}. The American power grid has more than 500,000 miles of high-voltage transmission lines and 5 million miles of distribution lines, and it is still rapidly growing with decarbonization initiatives \cite{ref3}. \textcolor{blue}{Inspecting such a large infrastructure is both labor-intensive and time-consuming, relying on traditional methods like manual patrols and helicopter surveys that have been in use for decades. Ground patrols allow thorough evaluations but are slow and costly, while helicopter surveys cover more areas quickly but with lower detection accuracy. Both approaches rely heavily on human visual observation, increasing the risk of missed or incorrect assessments \cite{ref4}}.
\color{blue}
\subsection{Vegetation Monitoring for Power Line Inspections}
\textcolor{blue}{Satellite imagery has been studied for vegetation monitoring within power line corridors. Sensors like Synthetic Aperture Radars (SAR) \cite{ref5}, operating in the microwave region of the electromagnetic spectrum, and optical sensors \cite{ref6}, capturing multispectral data in the visible and near-infrared regions, have been widely explored. Some studies \cite{ref7,ref8,ref9} have used multispectral stereo pairs of satellite images to identify trees with extracted Normalized Difference Vegetation Index (NDVI) information to evaluate encroachment to power lines; however, capturing stereo images is technically challenging and costly for large-scale monitoring \cite{ref10}. Recent studies focused on monocular satellite imagery combined with a machine learning technique, i.e., Adaboost classifier \cite{ref11} and Support Vector Machine (SVM) \cite{ref12} for extracting, segmenting, and classifying trees. However, these methods are constrained by data availability in low-light or cloudy conditions. Additionally, in these studies, the positions of power lines are typically predetermined, as they struggle to automatically detect power line conductors and bundles. Even the highest-resolution imagery is often too coarse to capture the thin and small features of power lines, particularly distribution lines, which are smaller, positioned lower, and frequently obscured by trees, making them unsuitable for fine-scale detection.} 
\color{black}
\IEEEpubidadjcol

\textcolor{blue}{Fixed vision sensors \cite{ref13} and ground-based vehicles \cite{ref14} have been explored for higher-resolution data acquisition, but their limited view angles present challenges. In contrast, airborne Light Detection and Ranging (LiDAR) three-dimensional (3D) modeling with higher flexibility, has gained greater attention in the industry, typically employing fixed-wing aircraft or helicopters. Utilities such as Pacific Gas and Electric (PG\&E) and Duke Energy have utilized LiDAR to assess risks near power lines \cite{ref15,ref16}. Existing studies \cite{ref17,ref18,ref19,ref20} showed that power line clustering, as well as the extraction and classification of trees, can be achieved by analyzing the 3D point cloud data using line detection (e.g., Hough transform) \cite{ref17}, intensity analysis \cite{ref18}, and advanced Deep Neural Networks (DNNs) (e.g., CA-PointNet++ \cite{ref19}, VEPL-Net \cite{ref20}, and RandLA-Net \cite{ref21}). Despite its effectiveness, the broader use of airborne LiDAR is limited by heavy payloads, high power consumption, and the need for high point density to map small objects \cite{ref22}. For instance, high-voltage transmission lines are easier to map due to elevated positions, whereas lower voltage distribution lines are often obscured by dense forests, making detection more challenging \cite{ref18}.}
\color{blue}
\subsection{Opportunities in UAV-Based Monitoring}
\color{black}
As an alternative, UAVs equipped with lightweight vision sensors offer a more cost-effective and practical solution. They can operate closer to the ground and capture high-resolution images in both the visible red, green, and blue (RGB) channels and infrared spectrums, providing detailed views of the environment \cite{ref23}. \textcolor{blue}{This makes it particularly effective for monitoring both transmission and distribution lines.} However, analyzing these captured images still relies on human observation, making it prone to errors and oversight. 


\begin{table*}[ht]
\centering
\color{blue} 
\caption{\textcolor{blue}{Overview of Different Data Resources and Techniques}}
\label{tab:evaluation_methods}
\setlength{\tabcolsep}{3pt} 
\renewcommand{\arraystretch}{1.2} 
\arrayrulecolor{blue} 

\begin{tabular}{|m{2.1cm}|m{4cm}|m{1.5cm}|m{1.5cm}|m{1.5cm}|m{2cm}|m{1.2cm}|m{2cm}|}
\hline
\textbf{Data Resource} 
& \textbf{Main Technique} 
& \textbf{Power Line Detection} 
& \textbf{Vegetation Detection} 
& \textbf{Data Resolution} 
& \textbf{Computational Cost} 
& \textbf{Position Accuracy} 
& \textbf{Communication Load} \\ 
\hline

\multirow{2}{*}{\parbox{2.1cm}{\centering Satellite SAR/\\Optical Imagery}}
& Stereo matching \& NDVI \cite{ref7,ref8,ref9}
& No
& Yes 
& Low 
& Medium
& Low 
& High 
\\ \cline{2-8}
& Adaboost, SVM \cite{ref11,ref12}
& No
& Yes 
& Medium
& High 
& Medium
& High  
\\ \hline

\multirow{2}{*}{\parbox{2.1cm}{\centering Airborne LiDAR\\ (fixed-wing/ helicopter)}}
& Hough Transform  \& Intensity analysis \cite{ref16,ref17}
& Yes 
& No 
& Medium
& Medium 
& Medium
& High 
\\ \cline{2-8}
& 3D point cloud data classification with DNNs \cite{ref18,ref19,ref20} 
& Yes 
& Yes
& Medium
& High 
& High 
& High 
\\ \hline

\multirow{4}{*}{\parbox{2.1cm}{\centering UAV-Based Cameras}}
& Image processing (Canny, Hough Transform, etc.) \cite{ref28,ref29} 
& Yes
& No 
& High 
& Low 
& Low 
& High
\\ \cline{2-8}
& CNN-based classification \cite{ref31,ref32}
& Yes
& No 
& High 
& Medium 
& Low
& High
\\ \cline{2-8}
& UNet-based segmentation \cite{ref33,ref34,ref35}
& Yes
& No 
& High 
& High 
& Medium 
& Medium 
\\ \cline{2-8}
& \textbf{Proposed PL-YOLO} 
& \textbf{Yes }
& \textbf{Yes }
& \textbf{High }
& \textbf{Medium }
& \textbf{High }
& \textbf{Low} 
\\ \hline

\end{tabular}
\color{black} 
\end{table*}

\textcolor{blue}{ Utility companies are beginning to pilot advanced wireless technologies, such as private Long Term Evolution (LTE) networks, to support intelligent sensors \cite{ref24}. However, it is still challenging to manage the communication and control of many inspection UAVs to support their transmission of a large volume of streaming image data. In addition, the critical control and non-payload data, including location and speed, demand robust communication that can be hindered by limitations in channel data rates. This can lead to data traffic congestion, potentially jeopardizing the safe navigation and operation of UAVs, which in turn poses risks to power line infrastructure and public safety.} Therefore, there is a critical need to develop automatic and real-time detection methods to enhance the efficiency and applicability of UAV inspections. 
\color{blue}
\subsection{Image-Based Detection of Power Lines}
\color{black}
Image-based automatic detection of power lines is challenging due to complex backgrounds and weak features \cite{ref25}. Prior works have mainly utilized traditional image processing methods, such as Hough or Radon transforms to detect line segments \cite{ref26,ref27}. These methods are often complemented with prior information or clustering algorithms to assemble the segments into complete power lines. Edge extraction algorithms, such as Canny, Line Segment Detector (LSD), and Edge Drawing Lines (EDLines) have also been studied for power line detection \cite{ref28,ref29}. However, the performance of these methods heavily depends on carefully calibrated threshold parameters and handcrafted rules, which often lack adaptability. Moreover, these techniques have limited robustness for distinguishing power lines from other linear structures in noisy aerial images.

Deep learning neural networks, especially CNNs, excel at recognizing and learning features for object detection in images using extensive parameters and deep layers. Their significant successes span various applications, from autonomous driving to medical imaging \cite{ref30}. Recent studies have explored the application of CNNs, for instance, VGG-19 and ResNet-50 for detecting power lines in aerial images \cite{ref31,ref32}. However, these approaches only classify images containing power lines rather than for detection and positioning. CNNs for object detection tasks mostly use Horizontal Bounding Boxes (HBBs). These boxes encompass a large space compared to the power lines with thin structures and varying orientations, thus cannot accurately position power lines \cite{ref33}. Therefore, some efforts focus on segmentation solutions that can detect power lines at the pixel level \cite{ref34,ref35,ref36}. These efforts involve using UNet \cite{ref37}, a U-shaped CNN architecture specialized in image segmentation and its variants based on different backbones, such as VGG-19 \cite{ref38} or GhostNet \cite{ref39}. However, pixel-level line segmentation is inherently complex, time-consuming, and thus not ideal for real-time application. The YOLO algorithm is a popular object detection model known for its speed and accuracy \cite{ref40}. First introduced in 2016, YOLO has iterated rapidly, including advancements in feature extraction, data augmentation, and model optimization.  YOLOv8 \cite{ref41}, introduced in 2023, brings enhancements to multiple tasks. Its Oriented Bounding Box (OBB) task has the potential to solve the power line detection challenge with its rotatable capability.

\begin{figure*}[t]
\centering
\includegraphics[width=0.9\linewidth]{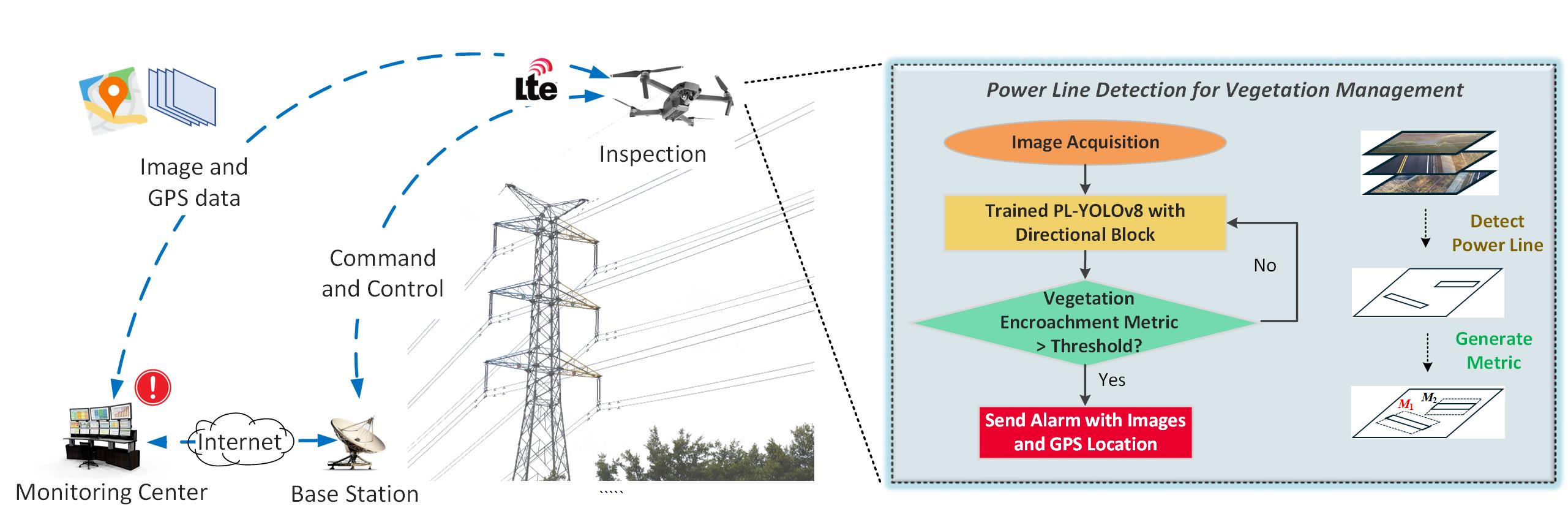}
\caption{Outline of PL-YOLO based power line detection for vegetation management.}
\label{fig_1}
\end{figure*}

This paper proposes an automatic real-time power line
detection framework based on UAV images. It can accurately
detect power lines with tight borders, quantitatively evaluate
nearby vegetation encroachment, and send location-based
alerts when risks are detected. The main contributions are listed as follows.
\begin{itemize}
    \item We introduce PowerLine-YOLOv8 (PL-YOLOv8), an enhanced version of the YOLOv8 algorithm, tailored for the specific challenges of power line detection. This model integrates a novel directional block composed of convolutional layers initialized with parameters and sizes derived from a directional filter bank. This bank uses specialized directional filters that are designed to detect the orientations and textures of power lines and their vicinity, enhancing the accuracy of OBBs for precise localization.
    \item PL-YOLOv8 represents the first adaptation of YOLOv8’s OBB capabilities specifically for power line extraction, resolving the trade-off between real-time detection and precise, tight bounding. For training, we developed a dataset with segmented image tiles and custom OBB annotations, specifically designed for accurate power line detection tasks. This dataset is customized using the Transmission Tower and Power Line Aerial (TTPLA) image dataset, which features high-resolution UAV imagery of power lines \cite{ref42,ref43}.
    \item We also propose a post-processing algorithm that develops a vegetation encroachment metric to evaluate the proximity of vegetation to power lines in aerial images. This metric utilizes the OBB detection locations and image data such as RGB indices, texture analysis, and brightness measurements. It offers an effective alternative to the intensive task of employing deep-learning-based segmentation methods for tree detection.  

    \item \textcolor{blue}{The proposed software framework is designed for deployment on UAV onboard processors. It transmits Global Positioning System (GPS) locations and essential image data only when vegetation metrics exceed predefined thresholds or when real-time viewing is requested. This location-based alert system minimizes data transmission, reduces wireless communication loads, and ensures high-quality image delivery. Its detailed, flexible views and low communication requirements make it ideal for monitoring both transmission and distribution lines.} 
    
\end{itemize}

\textcolor{blue}{The comparison of our proposed method with the existing literature is summarized in Table I.} The rest of the paper is organized as follows. Section II outlines the workflow of the proposed framework. Section III discusses the PL-YOLOv8 network, including the innovative directional block. Section IV details the vegetation encroachment post-processing algorithm. Section V shows the experimental results. Section VI provides the study conclusions.  
    
\section{Outline of Proposed Framework}

 \textcolor{blue}{Our proposed framework introduces an end-to-end solution to manage power line inspections and vegetation encroachment, as shown in Fig. \ref{fig_1}. Specifically, the software processes the image data as follows. }

 \textcolor{blue}{\textbf{\textit{Step 1}}: The UAV acquires images of the power lines while recording its GPS location. This data (images plus coordinates) is captured in real-time and can be stored locally on the UAV’s onboard processor.} 

 \textcolor{blue}{\textbf{\textit{Step 2}}: The onboard PL‑YOLOv8 algorithm analyzes these images to detect power lines, producing OBBs that accurately outline the detected lines. The PL-YOLOv8 is pretrained on the proposed dataset or supplemented with additional datasets as needed. } 

 \textcolor{blue}{\textbf{\textit{Step 3}}:  A vegetation encroachment metric module analyzes nearby pixels around the detected OBBs to evaluate how close or dense the vegetation is with respect to the power lines. It then computes an encroachment metric that quantifies potential risks. }

 \textcolor{blue}{\textbf{\textit{Step 4}}: The system checks the metrics against predefined thresholds. If any metric exceeds the acceptable limit, an alarm event is triggered. Upon an alarm, the UAV relays the relevant images and corresponding GPS coordinates to the monitoring center via the base station. Operators can thus access the critical data in near real-time, determining whether immediate action is required. }

 \textcolor{blue}{By embedding the PL‑YOLOv8 and metric calculation modules directly on the UAV, this workflow alleviates the communication load between the UAV, base station, and monitoring center while preserving image quality. If real-time viewing is needed, the system can additionally stream selected frames or sensor data over the wireless link. }

\section{Advanced PL-YOLOv8 for Power Line Detection}

Previous object detection algorithms, such as the original Region-based CNN (R-CNN) and its successors (Fast R-CNN, Faster R-CNN), typically employ a two-stage process \cite{ref44}. The localization stage employs a region proposal mechanism to pinpoint areas that potentially contain objects. This directs the subsequent neural network to concentrate on these areas, where it performs classification and regression tasks. This method achieves high accuracy through numerous regional proposals, but it also greatly limits time efficiency. YOLO revolutionizes real-time applications by employing an end-to-end neural network that simultaneously predicts bounding boxes and class probabilities. YOLOv8 builds on this by optimizing the architecture to improve object detection at various scales. It uses advanced convolutional layers and spatial pyramid pooling for better contextualization and spatial hierarchy preservation.

\begin{figure*}[t]
\centering
\includegraphics[width=0.95\linewidth]{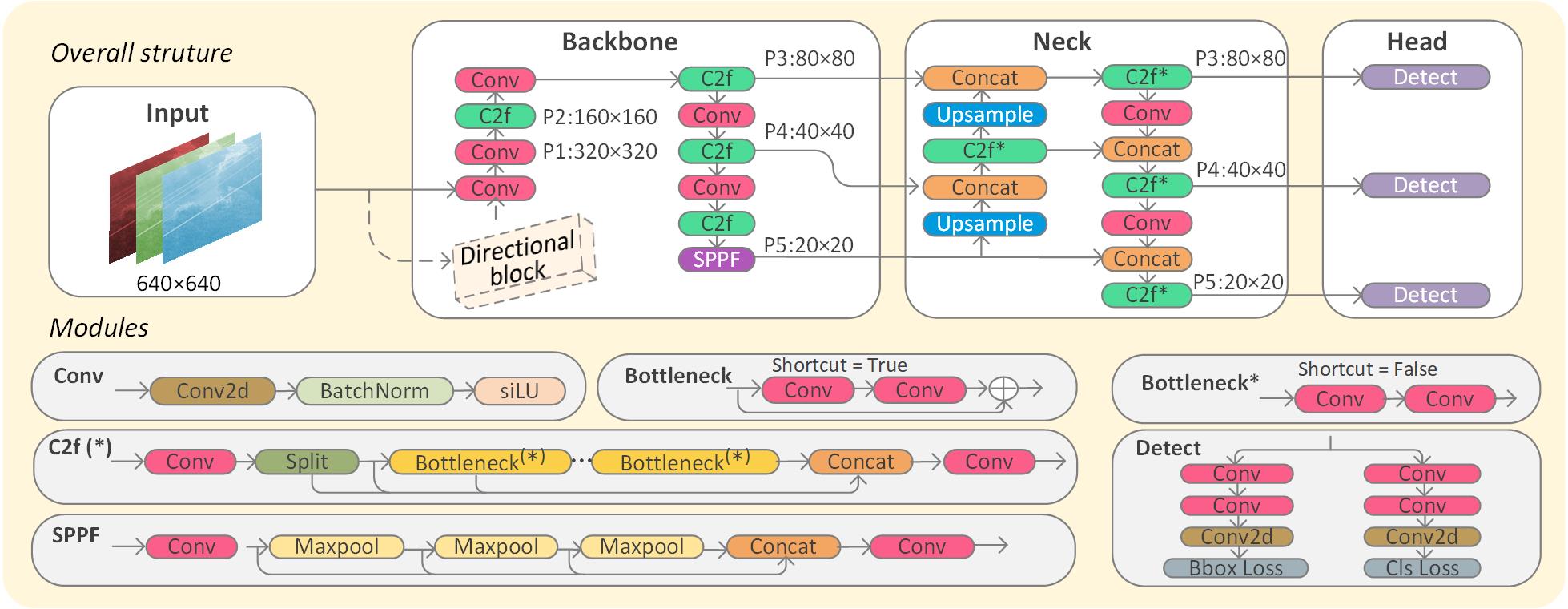}
\caption{Enhanced YOLO structure: PL-YOLOv8 with directional block.}
\label{fig_2}
\end{figure*}

\subsection{Overall structure}
Fig. \ref{fig_2} shows the PL-YOLOv8 network consisting of three main components: backbone, neck, and head. A directional block is incorporated at the beginning of the backbone, enhancing the standard YOLOv8. Further details on this block design are provided in section III(C). The backbone extracts image features using Conv, SPPF (Spatial Pyramid Pooling-Fast), and C2f modules \cite{ref45}. The Conv module includes a two-dimensional (2D) convolutional layer, batch normalization layer, and SiLU activation layer. SPPF can reduce computational load with three max-pooling layers that reduce the spatial dimensions of the feature maps. The C2f module enhances feature extraction using gradient diversion with Conv and residual modules (i.e., bottleneck). The neck employs a Feature Pyramid Network + Path Aggregation Network (FPN+PAN) structure to merge multi-scale features, enhancing detection accuracy for objects of varying sizes. The head uses a decoupled structure to optimize tasks of classification and detection separately, ensuring their respective optimal accuracy. In addition, the head adopts an anchor-free method by directly predicting bounding box coordinates from key points on the feature map. This eliminates the need for predefined anchor boxes in coupled heads, making the model more flexible and adaptive to various object sizes and shapes.

Given an example of a 640 × 640 image, YOLOv8 downsamples and generates feature maps (P1–P5) at scales of 320 × 320, 160 × 160, 80 × 80, 40 × 40, and 20 × 20 pixels, respectively. P1 and P2 have larger and shallower feature maps and retain detailed spatial information crucial for detecting small objects. P4 and P5, with smaller and deeper feature maps, capture abstract features necessary for recognizing large objects. By integrating the P3, P4, and P5 layers, YOLOv8 leverages the strengths of multiple scales to detect objects of various sizes.

\begin{figure}[!t]
\centering
    \hspace*{-7mm}
\subfloat[]{\includegraphics[width=0.48\linewidth]{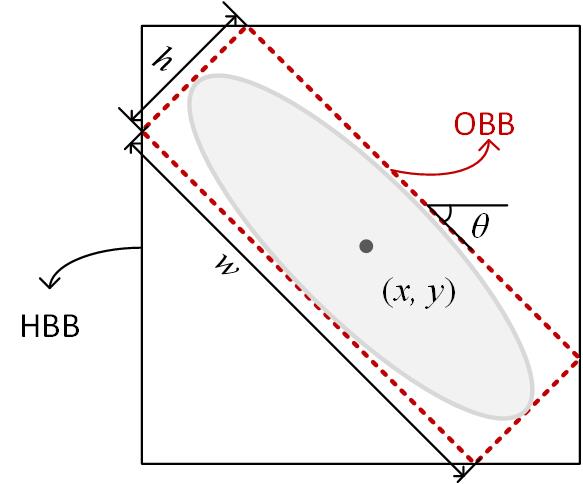}%
\label{fig_first_case}}
\hfil
\subfloat[]{\includegraphics[width=0.38\linewidth]{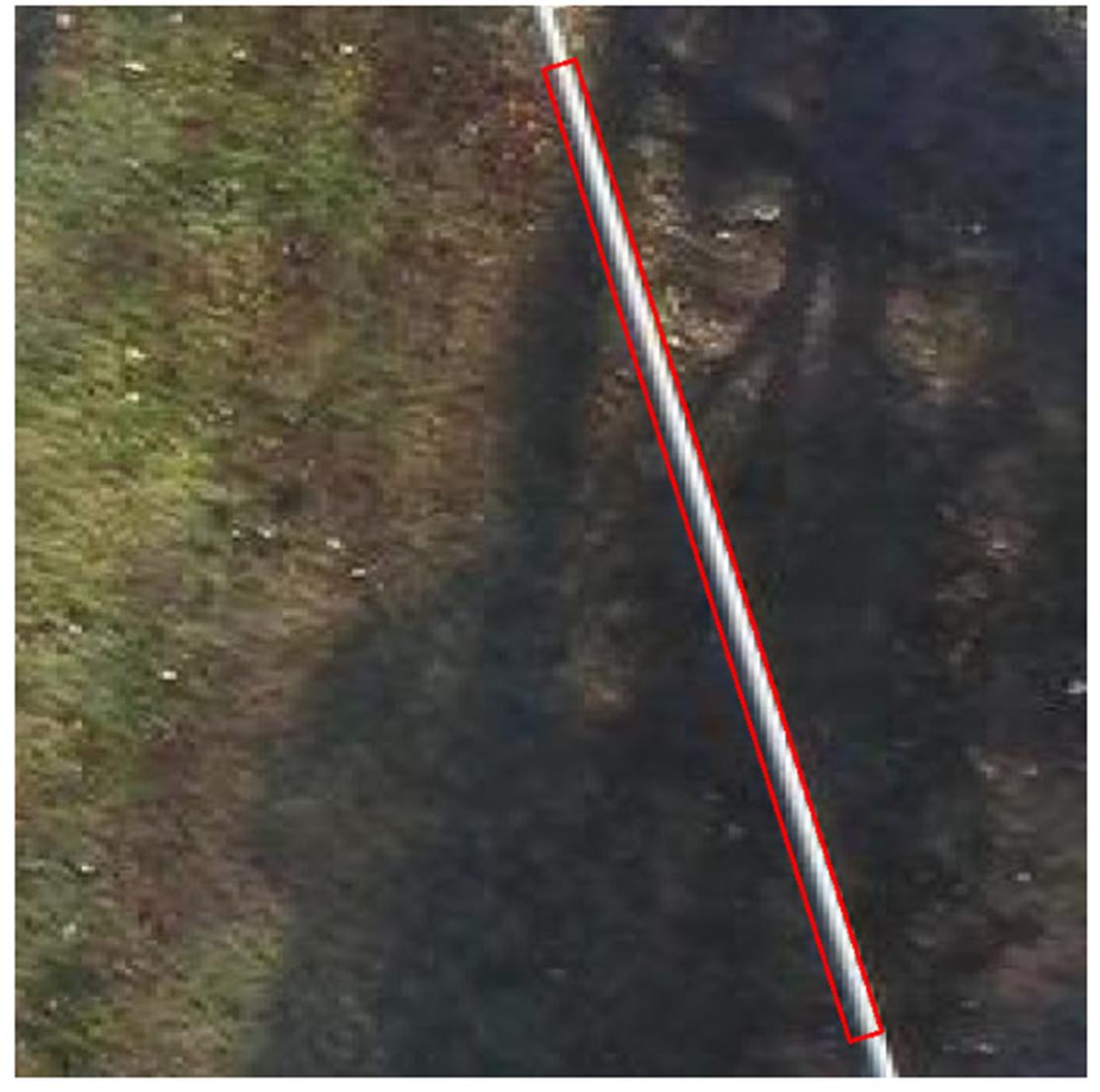}%
\label{fig_second_case}}
\caption{OBB task: (a) OBB representation. (b) Power line detection example using the OBB concept.}
\label{fig_3}
\end{figure}
\subsection{OBB task head}
The OBB task improves traditional HBB detection by considering the object's orientation, which is particularly useful for power line detection, as shown in Fig. \ref{fig_3}. An OBB is defined by its center coordinates \((x, y)\), width \(w\), height \(h\), and rotation angle \(\theta\), allowing for tighter bounds and reducing background inclusion. The OBB localization loss is computed using the Probability-based Intersection over Union (ProbIoU) method \cite{ref46} instead of the traditional IoU loss for HBBs. This method encodes the parameters (\(x\), \(y\), \(w\), \(h\), \(\theta\)) into a covariance matrix and calculates their various distance metrics, ensuring accurate localization loss by considering the object's position, size, and rotation.

\subsection{Directional block design}

\begin{figure*}[t]
\centering
\includegraphics[width=0.95\linewidth]{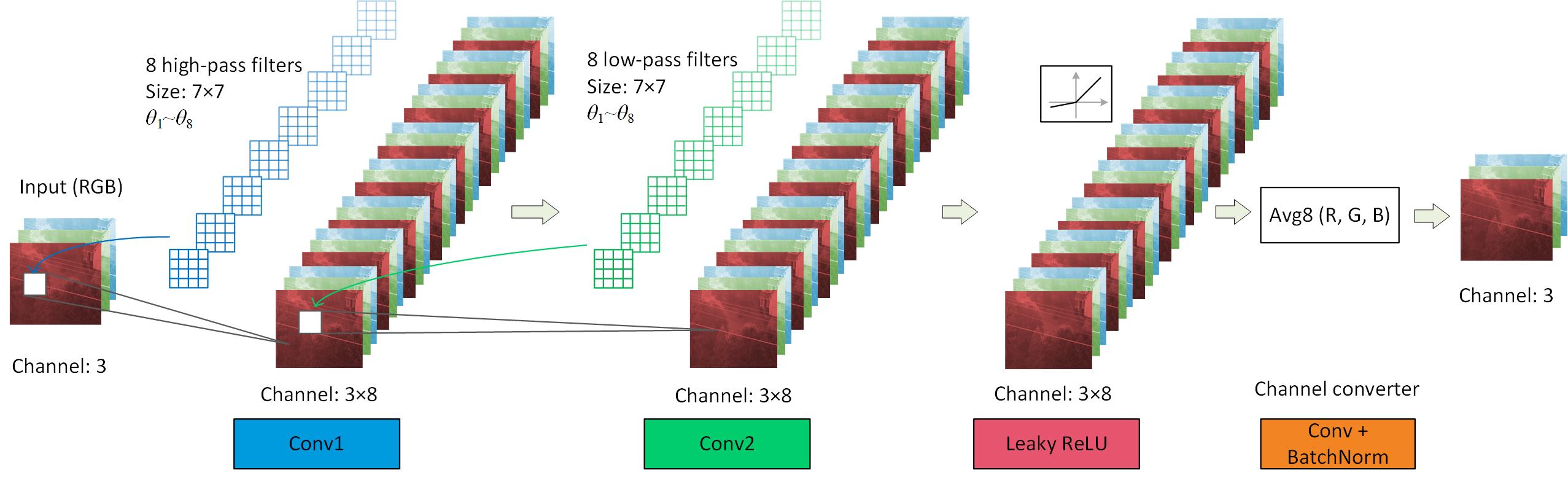}
\caption{Directional block containing 8 directional high-pass filters in its first stage.}
\label{fig_4}
\end{figure*}
\begin{figure}[!t]
\centering
    \hspace*{-5mm}
\subfloat[]{\includegraphics[width=0.48\linewidth]{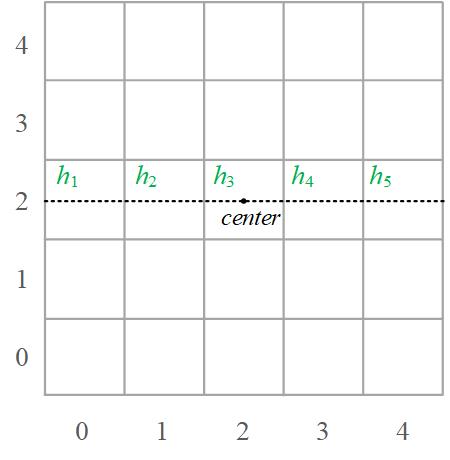}%
\label{fig5_first_case}}
\hfil
\subfloat[]{\includegraphics[width=0.48\linewidth]{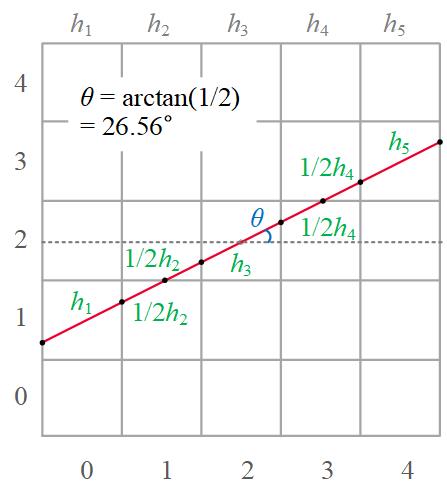}%
\label{fig5_second_case}}
\caption{2D filter rotation example when $n=5$. (a) Filter $f_{\theta=0^\circ}(i,j)$. (b) Filter $f_{\theta=25.56^\circ}(i,j)$.}
\label{fig_5}
\end{figure}

The directional block is designed to comprise two convolutional layers, a leaky ReLU activation layer, and a channel converter, as illustrated in Fig. \ref{fig_4}. The leaky ReLU activation helps mitigate negative values, preventing inactive neurons during training. The channel converter consists of a convolutional layer and a batch normalization layer. The convolutional layer features a \(1 \times 1\) kernel that averages directional features from 8 channels and converts them to 3 RGB channels. This configuration enables seamless integration into the YOLO backbone for further processing. The parameters in the convolutional layers are initialized using directional filters, as designed below.
\begin{figure}[!t]
\centering
{\includegraphics[width=0.95\linewidth]{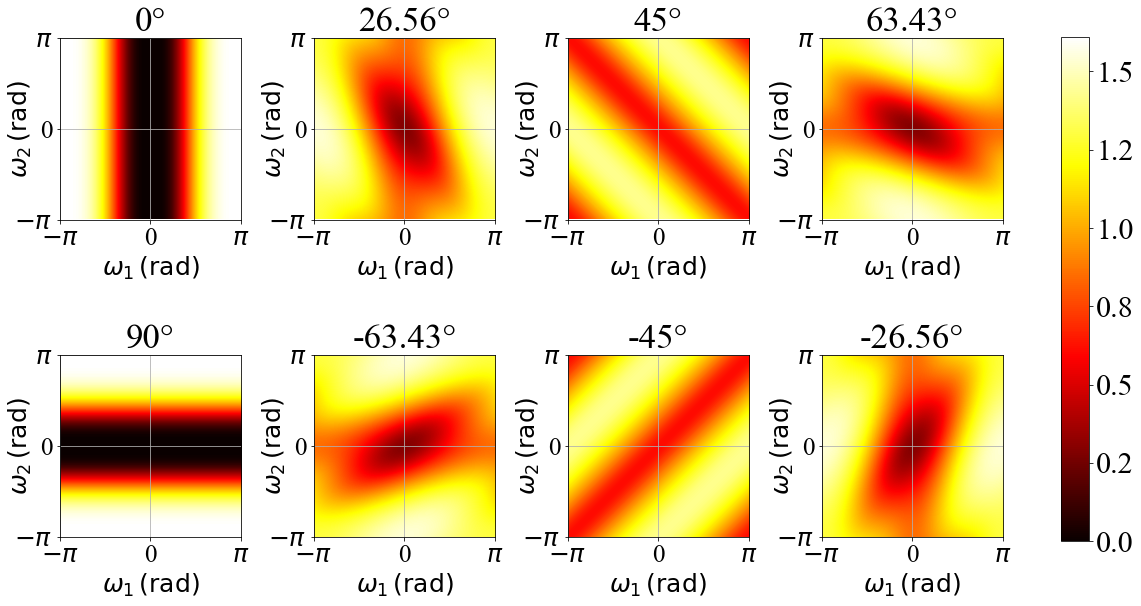}}%

\caption{2D Frequency response plots, $|H_{bp}(e^{j(\omega_1,\omega_2)})|$ for $(\omega_1,\omega_2) \in (-\pi, \pi)$ at $\theta = \{0^\circ, 26.56^\circ, 45^\circ, 63.44^\circ, 90^\circ, -63.44^\circ, -45^\circ, -26.56^\circ\}$.}

\label{fig_6}
\end{figure}
In image processing, it is customary to apply a low-pass filter to remove the noise while detecting the edges using an edge detection filter, which is essentially a high-pass filter. This approach generates a wavelet band-pass filter, as shown in (\ref{deqn_1}).

\begin{equation}
\label{deqn_1}
    h_{bp}[n] = (h_{lp}*h_{hp})[n]
\end{equation}
Here, $h_{bp}$ represents the band-pass filter. It is noted that the
frequency response $H_{bp}(e^{j\omega})$ of this filter is not identically
zero across all frequencies $\omega$ because the high- and low-pass
filters are inherently not perfect. This results in a practical
band-pass filter that does not entirely block all frequencies.

We rotate the one-dimensional (1D) prototype filter $h_{bp}$
through eight directions within the range $(\omega_1,\omega_2) \in (-\pi, \pi)$, as described in the following paragraph and shown in Fig. \ref{fig_5}.
This yields a set of directional filters, with their smooth
frequency responses shown in Fig. \ref{fig_6}. These filters are designed
to emphasize edges in all eight directions. This feature is vital
for highlighting weak, straight-edge features of power lines against complex backgrounds, which is crucial for accurate
detection and analysis. \textcolor{blue}{The designed filter parameters can initialize directional blocks in deep learning models to enhance directional feature extraction, allowing deeper layers in capturing structural details. This approach ensures that the network efficiently prioritizes essential image attributes from the start, improving the robustness of power line detection.}

As pointed out above, the directional filters are created by rotating a 1D prototype filter with an impulse response of \(f_h=\{h_1, h_2,..., h_n\}\) into a 2D space across various angles parameterized by \(\theta\). Illustrated in Fig. \ref{fig_5}, we use a \(n \times n\) grid to represent the 2D filter \(f_\theta (i,j)\), where \(i\) and \(j\) are the horizontal and vertical indices of the grid cells, respectively. A line at angle \(\theta\) crosses the grid's center with coordinates of \((\frac{n}{2},\frac{n}{2})\) and is defined by equation (\ref{deqn_2}):
\begin{equation}
\label{deqn_2}
    y = (x - \frac{n}{2}) \times \tan\theta + \frac{n}{2}
\end{equation}

Specifically, the line segments in each grid denote the weights of the filter parameter. When \(\theta= 0\)  and \(n\) is odd, \(f_0(i,\frac{n-1}{2})\) equals the original filter without rotation. Otherwise, the values in \(f_\theta(i,j)\) are in proportion to the line segment length \(L_{\theta}(i,j)\), and are normalized by the line length in the center as \(L_{\theta}(\frac{n-1}{2},\frac{n-1}{2})\), as shown in (\ref{deqn_3}).
\begin{equation}
\label{deqn_3}
    f_\theta(i,j)=f_h(i)\times \frac{L_{\theta}(i,j)}{L_{\theta}(\frac{n-1}{2},\frac{n-1}{2})}
\end{equation}
where \(L_{\theta}(\frac{n-1}{2},\frac{n-1}{2})=\frac{1}{\cos \theta}\) when \(\theta \leq  45^\circ \).  \(L_{\theta}(i,j)\) is calculated by finding out the entry and exiting points in cell \((i,j)\), and calculating their Euclidean distances. The calculation algorithm is developed as follows. 
\begin{algorithm}[H]
\caption{\(L_\theta(i,j)\) Calculation}
\label{alg:lij_calculation}
\begin{algorithmic}
    
\STATE {\textsc{CALCULATE\_LINE\_LENGTHS}}$(n, \theta)$
\STATE  \hspace{0.3cm} $center \gets n / 2$
\STATE  \hspace{0.3cm} \textbf{for} each cell $(i, j)$ in $n \times n$ grid:
\STATE  \hspace{0.8cm} \textbf{determine}\text{ coordinates of four cell corners:}
\STATE  \hspace{0.8cm} $(\mathbf{X_r, Y_r})_{i,j} \gets (i + r, j + r)$ \textbf{for} $r = 0, 1$
\STATE  \hspace{0.8cm} \textbf{determine} \text{intersection points on cell edges:}
\STATE  \hspace{0.8cm} $\mathbf{X_{\text{at } Y_r}} \gets {(y - center)}/{\tan \theta} + center$ \textbf{for} $y \in \mathbf{Y_r}$
\STATE  \hspace{0.8cm} $\mathbf{Y_{\text{at } X_r}} \gets (x - center) \times \tan \theta + center$ \textbf{for} $x \in \mathbf{X_r}$
\STATE  \hspace{0.8cm} \textbf{identify} \text{ two valid intersection points in the cell:}
\STATE  \hspace{0.8cm} \textbf{if} $X_{0} < X_{\text{at} Y_r} < X_{1}$ \text OR \STATE  \hspace{1.2cm}  $Y_{0} < Y_{\text{at} X_r} < Y_{1}$ \textbf{for} $r = 0, 1$ \text{is true}:
\STATE  \hspace{1cm}$\{(x_1,y_1),(x_2,y_2)\} \gets \{(X_{\text{at} Y_r}, Y_r),( X_r, Y_{\text{at} X_r})\}$ 

\STATE  \hspace{0.8cm} \textbf{Length}: $ L_\theta(i,j) \gets \sqrt{(x_2 - x_1)^2 + (y_2 - y_1)^2}$
\STATE  \hspace{0.3cm} \textbf{return}  $L_\theta(i,j)$

\end{algorithmic}
\end{algorithm}

For angles \(\theta \geq 45^\circ\), the filter is generated by transposing \(f_{(90^\circ-\theta)}\), which is calculated using the Algorithm 1. Within the directional block, the angles are set at \(\theta=\{0^\circ,  \pm26.56^\circ, \pm45^\circ, \pm63.44^\circ, 90^\circ\}\). The initial parameters for the first two convolutional layers are created by rotating a high-pass filter and a low-pass filter, respectively. \textcolor{blue}{We selected the 1D high-pass prototype filter obtained from a seven-order half-band Lagrangian maximally flat low-pass filter with the following transfer function (\ref{deqn_4}).}
\color{blue}
\begin{equation}
\label{deqn_4}
\begin{split}
   H_{hp}(z) = 1 + \frac{9}{16} \left((-z)^1 + (-z)^{-1}\right) \\
         - \frac{1}{16} \left((-z)^3 + (-z)^{-3}\right)
\end{split}
\end{equation}

The high-pass filter is obtained using the transformation $ H_{hp}(z) = H_{lp}(-z)$, and the corresponding impulse response of the high-pass filter is given by (\ref{deqn_5}). 
\begin{equation}
\label{deqn_5}
h_{hp}[n]=\{\frac{1}{16},0,-\frac{9}{16},1,-\frac{9}{16},0,\frac{1}{16}\} 
\end{equation}

\textcolor{blue}{The low-pass filter used in the directional block is a wide-band low-pass filter, empirically designed. Its impulse response is shown in by (\ref{deqn_6}). }
\begin{equation}
\label{deqn_6}
   h_{lp}[n]=\{-\frac{1}{4},-\frac{1}{2},1,2,1,-\frac{1}{2},-\frac{1}{4}\}
\end{equation}

\begin{figure}[!t]
\centering
\subfloat[]{\includegraphics[width=0.48\linewidth]{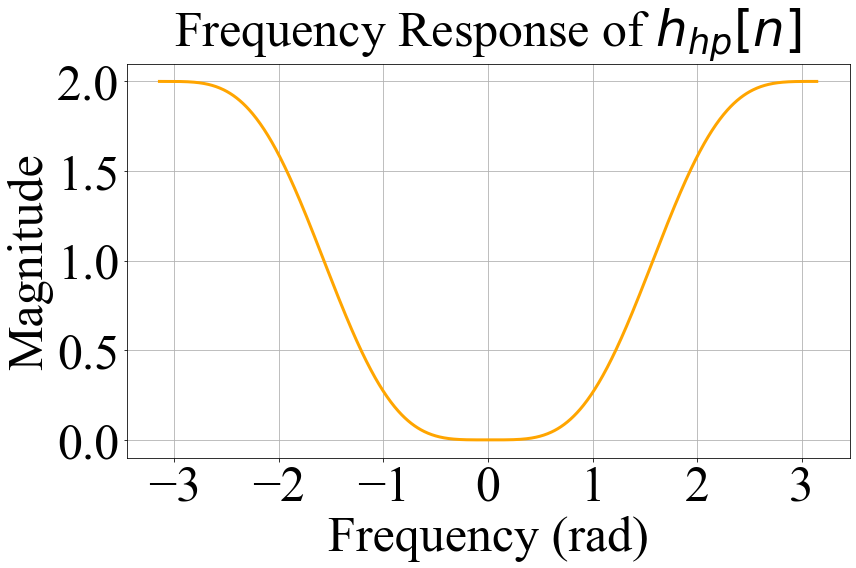}%
\label{fig7_first_case}}
\hfil
\subfloat[]{\includegraphics[width=0.48\linewidth]{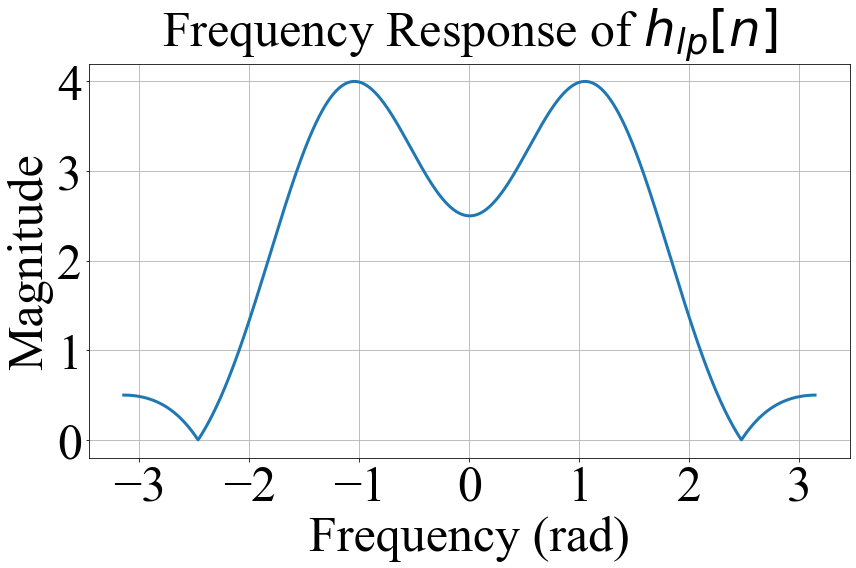}%
\label{fig7_second_case}}
\caption{Frequency responses of (a) $h_{hp}[n]$ and (b) $h_{lp}[n]$, respectively.}
\label{fig_7}
\end{figure}

The corresponding frequency responses within $(-\pi, \pi)$ are shown in Fig. \ref{fig7_first_case} and Fig. \ref{fig7_second_case}, respectively. Fig. \ref{fig7_first_case} illustrates that the high frequencies are preserved while the low frequencies are suppressed, making the filter effective for edge detection. Fig. \ref{fig7_second_case} illustrates that the low-pass filter, designed as a wide-band filter, suppresses high-frequency noise while preserving low frequencies and maintaining essential structural details. Additionally, neither filter is ideal, allowing nonzero values within the suppressed frequency bands to ensure that critical information is retained.

\color{black}
The PL-YOLOv8 model with the initial directional block achieves a higher accuracy than standard PL-YOLOv8 in detecting power lines using OBBs.

\section{Vegetation Encroachment Metric Development} 

Using the location results from power line detection with PL-YOLOv8, we devise a method to measure vegetation encroachment that assesses the density and growth of vegetation near power lines in the images. This approach combines a Greenness Index (GI) with a Tree-Grass Differentiation Index (TGDI) to create a comprehensive vegetation encroachment metric. Deep learning-based segmentation solutions are widely discussed and utilized for pixel-level vegetation detection \cite{ref48}.  However, they encounter significant computational challenges, especially for the detection of trees with diverse morphology. Our method employs RGB indices and image texture and brightness data, which can tackle the computational challenges associated with segmentation tasks in neural networks for tree detection.

\subsection{Greenness index}

The GI is derived from the Green Red Vegetation Index (GRVI), which measures image greenness, complemented by a Gaussian filter that summarizes spatial information. GRVI is frequently utilized in image-based vegetation analysis and is represented in (\ref{deqn_7}).

\begin{equation}
\label{deqn_7}
 \text{GRVI} = \frac{G - R}{G + R}
\end{equation}
where \( G \) and \( R \) are the green and red channel pixel values, respectively. This index effectively highlights vegetation by measuring the relative intensity of green compared to red light. 

\begin{figure}[!t]
\centering
\includegraphics[width=0.95\linewidth]{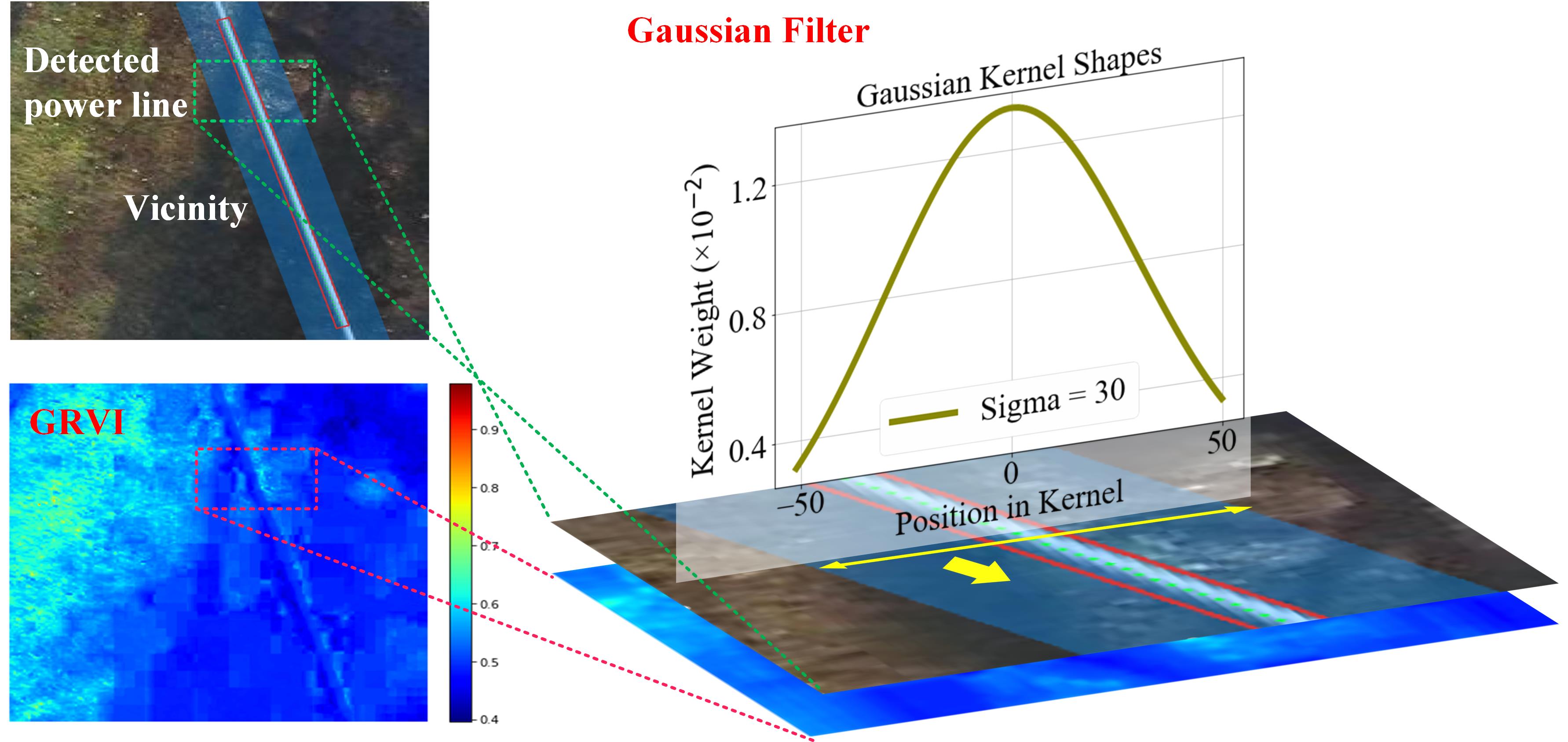}
\caption{Greenness index generation example.}
\label{fig_8}
\end{figure}

To aggregate the greenness information near power lines, we use a discrete, normalized Gaussian filter as outlined in (\ref{deqn_8}).

\begin{equation}
\label{deqn_8}
   G(g) = \frac{\exp\left(-0.5 \left(\frac{g}{\sigma}\right)^2\right)}{\sum_{g} \exp\left(-0.5 \left(\frac{g}{\sigma}\right)^2\right)}
\end{equation}
Here, $g$ represents the index, represented by the set $\{-\frac{z-1}{2}, \ldots, 0, \ldots, \frac{z-1}{2}\}$, where $z$ is the filter size and $\sigma$ is the standard deviation, influencing the filter's spread.

Fig. \ref{fig_8} illustrates the application of the Gaussian filter near the power line. The power line's OBB is highlighted in red and defined by \(\{(u_c, v_c), (w, h), \theta\}\). Here, \( (u_c, v_c) \) are the center coordinates, \( w \) and \( h \) the width and height, respectively, and \( \theta \) the rotation angle. The central line of the OBB is described by (\ref{deqn_9}).
\begin{equation}
\label{deqn_9}
(u, v) = (u_c, v_c) + t \cdot (\cos\theta, \sin\theta)
\end{equation}
where \(t\) varies from \(-\frac{w}{2}\) to \(\frac{w}{2}\). Along this line, \(m\) evenly spaced pixels are sampled and defined in (\ref{deqn_10}).

\begin{equation}
\label{deqn_10}
\begin{aligned}
(u_k, v_k) = (u_c &+ (-\frac{w}{2} + k \cdot \frac{w}{m}) \cos\theta,\\ 
                  & v_c + (-\frac{w}{2} + k \cdot \frac{w}{m}) \sin\theta)
\end{aligned}
\end{equation}
where $k$ ranges from $0$ to $m$. \textcolor{blue}{The value of \( m \) is chosen as 100, considering the image size in our experiment is \( 640 \times 640 \).} 

At each pixel sample $(u_k, v_k)$, the 1D Gaussian filter is centered and applied perpendicular to the central line. Let \( \mathbf{p}_\perp \) be the unit vector perpendicular to the line's direction at the point \( (x, y) \). If the direction vector of the line is \( \mathbf{d} = (dx, dy) \), then \( \mathbf{p}_\perp = (-dy, dx) \). Points along the line perpendicular to \( \mathbf{d} \) at \( (u_k, v_k) \) are determined by (\ref{deqn_11}).
 \begin{equation}
\label{deqn_11}
     (u_\perp, v_\perp)_g = (u_k, v_k) + g \cdot \mathbf{p}_\perp
\end{equation}
Here, \( g \) represents the index in the Gaussian kernel, and \( \mathbf{p}_\perp \) is normalized for uniform spacing.

Next, the weighted greenness value \( W(k) \) for each position \((u_k, v_k)\) along the power line is calculated as a weighted sum of the GRVI values using Gaussian weights, as shown in (\ref{deqn_12}).

\begin{equation}
\label{deqn_12}
     W(k)= \sum_{g=-\frac{n-1}{2}}^{\frac{n-1}{2}} I(u_\perp, v_\perp)_g  \cdot G(g)
\end{equation}
Here, $I(u_\perp, v_\perp)_g$ denotes the GRVI value at each perpendicular pixel indexed by $g$, and $G(g)$ is the corresponding Gaussian kernel weight. This results in a set of $\mathbf{W(k)}$ values for the power line. Finally, a composite greenness index \( GI \) is derived by considering both the maximum and the average of these values, as shown in (\ref{deqn_13}). 
\begin{equation}
\label{deqn_13}
GI = \max(\mathbf{W(k)}) + \text{mean}(\mathbf{W(k)})
\end{equation}

The GI is specifically designed to provide a comprehensive evaluation of vegetation and its proximity to power lines. This index effectively captures both extreme and typical greenness scenarios near the power lines, facilitating a robust assessment of vegetation encroachment. Additionally, it employs a Gaussian filter to process spatial information, prioritizing areas closer to the power line with higher weights while assigning lower weights to farther regions. As a result, the GI provides a reliable measure for monitoring vegetation health and density around power lines.

\subsection{Vegetation encroachment metric}
The GI provides color-based information from images, which can make it challenging to distinguish between highly green grasslands and trees. To address this issue, we incorporate the TGDI \cite{ref49}, which utilizes the visual distinctions between grasslands and tree crowns; grass typically appears brighter and smoother, while tree crowns are characterized by their darker and more textured appearance. The TGDI is defined as follows in (\ref{deqn_14}).
\begin{equation}
\label{deqn_14}
TGDI = \log_{10}(T) \times B
\end{equation}

Here, \( T \) denotes the proportion of pixels identified as edges by the Canny algorithm, reflecting the image's texture. \( B \) represents the average grayscale values of the pixels, indicating the image's brightness. Since brightness is a more significant factor than texture in differentiating grasslands from tree crowns, the texture component \( T \) is adjusted by applying a logarithmic transformation, serving as an adjustment coefficient in the index. In this formulation, a higher TGDI value indicates the presence of grass, while a lower value suggests the presence of trees.

Combining this with the GI, we derive a final metric for vegetation encroachment, as represented in (\ref{deqn_15}).
\begin{equation}
\label{deqn_15}
M = \alpha \cdot GI + \beta \cdot TGDI
\end{equation}
Here, \( \alpha \) and \( \beta \) are coefficients that calibrate the contribution of each index to the overall metric, ensuring a balanced assessment of vegetation presence. Notably, \( \beta \) is set to a negative value to ensure that a higher \( M \) indicates a higher likelihood of tree presence, aligning with the intended interpretation of the metric.

\section{Experiment Results} 
\textcolor{blue}{The proposed framework was developed and demonstrated using Python 3.10, leveraging its library support for deep learning and computer vision applications. Model design and implementation were carried out using PyTorch, with Spyder as the development environment and Anaconda for environment management.} The effectiveness of the proposed PL-YOLOv8 and vegetation encroachment metric is demonstrated using a comprehensive public power infrastructure dataset, TTPLA \cite{ref42, ref43}. This dataset comprises UAV low-altitude aerial images of transmission towers and power lines collected from two states in the U.S. We preprocessed the dataset into appropriate image sizes and generated corresponding OBB annotations. The dataset preprocessing, detection results, evaluation, and assessment are discussed as follows.

\begin{table*}[t]
\caption{Comparison of Object Detection Models\label{table1}}
\centering
\label{table2}
\begin{tabular}{|c||c|c|c||c|c|c|}
\hline
\hspace{0.6cm}\textbf{Model}\hspace{0.6cm} & \hspace{0.4cm}\textbf{Image Size}\hspace{0.4cm} & \hspace{0.5cm}\textbf{Dataset}\hspace{0.5cm} & \hspace{0.5cm}\textbf{Task}\hspace{0.5cm} & 
\hspace{0.7cm} $\mathbf{AP^{50}}$ \hspace{0.7cm} & 
\hspace{0.5cm} $\mathbf{AP^{50-90}}$ \hspace{0.5cm} &
\hspace{0.5cm}\textbf{FPS}\hspace{0.5cm}  \\ 
\hline
\textcolor{blue}{\multirow{2}{*}{Yolact++ (ResNet50)}} & \textcolor{blue}{\multirow{2}{*}{$700 \times 700$}} & \textcolor{blue}{\multirow{2}{*}{Original TTPLA}} & \textcolor{blue}{Mask} & \textcolor{blue}{$5.11\%$} & \textcolor{blue}{$1.64\%$} & \textcolor{blue}{\multirow{2}{*}{9.04}} \\
\cline{4-6}
&  &  & \textcolor{blue}{HBB} & \textcolor{blue}{$41.45\%$} & \textcolor{blue}{$22.29\%$} & \\
\hline
\textcolor{blue}{\multirow{2}{*}{Yolact++ (ResNet101)}} & \textcolor{blue}{\multirow{2}{*}{$700 \times 700$}} & \textcolor{blue}{\multirow{2}{*}{Original TTPLA}} & \textcolor{blue}{Mask} & \textcolor{blue}{$5.38\%$} & \textcolor{blue}{$1.92\%$} & \textcolor{blue}{\multirow{2}{*}{9.72}} \\
\cline{4-6}
&  &  & \textcolor{blue}{HBB} & \textcolor{blue}{$42.32\%$} & \textcolor{blue}{$22.37\%$} & \\
\hline
\textcolor{blue}{YOLOv8(\textit{n})} & \textcolor{blue}{\multirow{9}{*}{$640 \times 640$}} & \textcolor{blue}{\multirow{9}{*}{TTPLA-Tile-OBB}} & \textcolor{blue}{\multirow{9}{*}{OBB}} & \textcolor{blue}{$69.00\%$} & \textcolor{blue}{$36.02\%$} & \textcolor{blue}{$345$}\\
\cline{1-1} \cline{5-6} \cline{7-7}
\textcolor{blue}{\textbf{PL-YOLOv8(\textit{n})}} &  &  &  & \textcolor{blue}{$\mathbf{78.24\%}$} & \textcolor{blue}{$\mathbf{43.50\%}$} & \textcolor{blue}{$\mathbf{97}$}\\
\cline{1-1} \cline{5-6} \cline{7-7}
YOLOv8(\textit{s}) &  &  &  & $76.33\%$ & $41.14\%$ & $151.51$\\
\cline{1-1} \cline{5-6} \cline{7-7}
\textbf{PL-YOLOv8(\textit{s})} &  &  &  & $\mathbf{78.14\%}$ & $\mathbf{44.32\%}$ & $\mathbf{49.50}$\\
\cline{1-1} \cline{5-6} \cline{7-7}
YOLOv8(\textit{m}) &  &  &  & $70.70\%$ & $40.50\%$ & $60.24$\\
\cline{1-1} \cline{5-6} \cline{7-7}
\textbf{PL-YOLOv8(\textit{m})} &  &  &  & $\mathbf{71.94\%}$ & $\mathbf{41.14\%}$ & $\mathbf{20.37}$\\
\cline{1-1} \cline{5-6} \cline{7-7}
YOLOv8(\textit{l}) &  &  &  & $73.22\%$ & $41.39\%$ &$35.08$\\
\cline{1-1} \cline{5-6} \cline{7-7}
\textbf{PL-YOLOv8(\textit{l})} &  &  &  & $\mathbf{74.60\%}$ & $\mathbf{43.38\%}$ &$\mathbf{16.47}$\\
\hline
\end{tabular}
\end{table*} 

\begin{figure}[!t]
\centering

\includegraphics[width=0.95\linewidth]{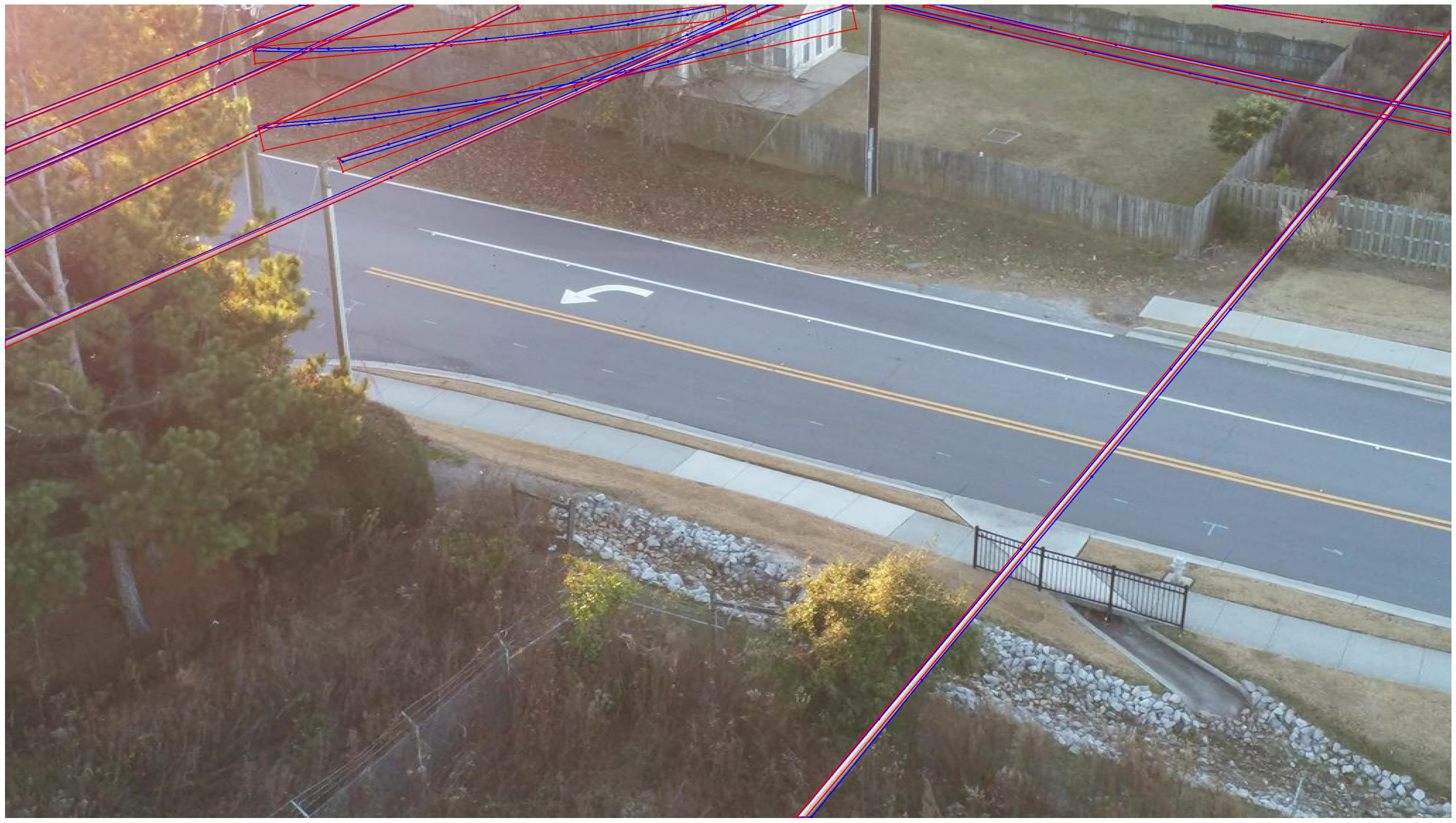}
\caption{OBB annotation generation in an original image.}
\label{fig_9}
\end{figure}

\begin{figure}[!t]
\centering
\includegraphics[width=0.95\linewidth]{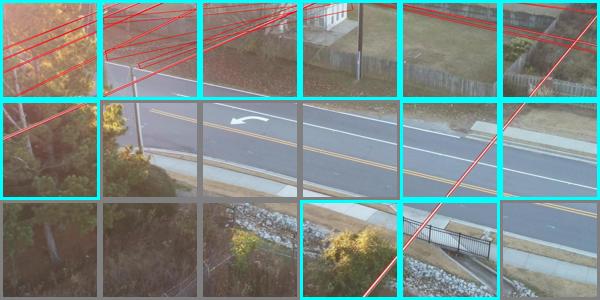}
\caption{Image tiles containing OBB annotations.}
\label{fig_10}
\end{figure}

\subsection{Data preprocessing}
\textcolor{blue}{The TTPLA dataset consists of 1,242 images with a resolution of $3,840\times2,160$ pixels and includes manually labeled 10,075 instances of power lines.} It can be utilized for the detection and segmentation tasks. For our purposes, we used only the power line data. We developed an automatic labeling method that converts segmentation annotations to OBB annotations. The segmentation annotations are polygons formed by a set of points along the instance edges. We used the \textit{convexHull} and \textit{minAreaRect} functions in OpenCV to generate OBBs that enclose the polygons. Fig. \ref{fig_9} illustrates the original polygon annotations (in blue) and the newly generated OBB annotations (in red). These OBB annotations efficiently enclose the power line within the minimal rectangular area, showcasing their effectiveness.

\begin{table}[!t]
\color{blue}
\caption{YOLOv8 OBB Model Sizes}
\label{table3}
\centering
\begin{tabular}{|c|c|c|c|}
\hline
\textbf{Model Size} & \textbf{Depth Factor \textit{d}} & \textbf{Width Factor \textit{w}} & \textbf{Params (M)} \\ \hline
\textit{n}        & 0.33             & 0.25              & 3.1                   \\ \hline
\textit{s}        & 0.33             & 0.50              & 11.4                   \\ \hline
\textit{m}         & 0.67              & 0.75             & 26.4                  \\ \hline
\textit{l}          & 1              & 1             & 44.5                   \\ \hline
\end{tabular}

\end{table}
\color{black}

Furthermore, we divided the original images into $640\times640$ tiles/sub-images. This ensures that the power lines in images are processed thoroughly, preserving important details without substantial loss during resizing. The OBB annotations were generated for each tile by adjusting and aligning with the boundaries. Fig. \ref{fig_10} highlights tiles containing power lines, which are preserved in the curated dataset. This dataset, named TTPLA-Tile-OBB, now comprises 17,178 images of size $640\times640$ pixels, with accurate OBB annotations for power lines. It was then divided into training, validation, and test sets with an 8:1:1 ratio.

\subsection{Power line detection results}

We conducted various training scenarios using different datasets and models to evaluate performance. The original and preprocessed TTPLA dataset are used to train both the standard YOLOv8 model and the proposed PL-YOLOv8 across four different model sizes: nano (\textit{n}), small (\textit{s}), medium (\textit{m}), and large (\textit{l}), respectively. 
\textcolor{blue} {They are scaled by the depth and width factors (\textit{d} and \textit{w}), which determine the number of layers (depth) and the number of channels per layer (width) in the network, thereby resulting in different numbers of parameters, as shown in TABLE \ref{table3}.}

\begin{figure*}[t]
\centering
\subfloat[]{\includegraphics[width=0.75\linewidth]{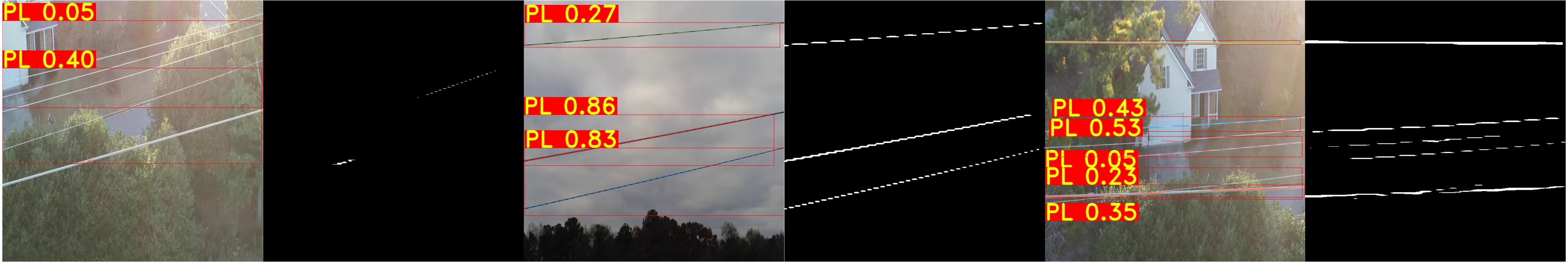}%
\label{fig11_first_case}}
\hfil
\hspace*{-3mm}
\subfloat[]{\includegraphics[width=0.48\linewidth]{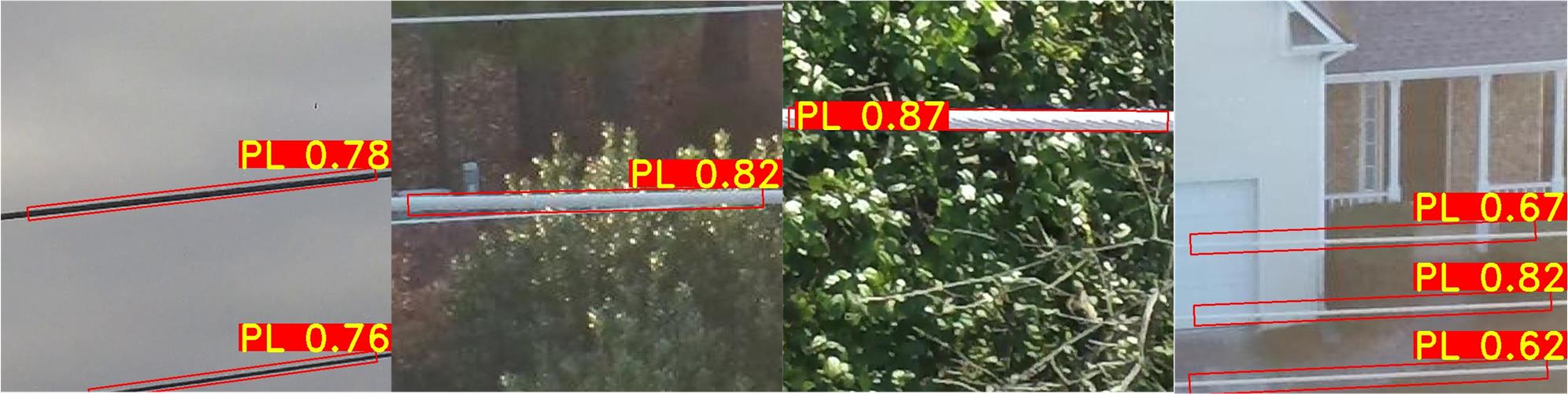}%
\label{fig11_second_case}}
\hfil
\hspace*{3mm}
\subfloat[]{\includegraphics[width=0.48\linewidth]{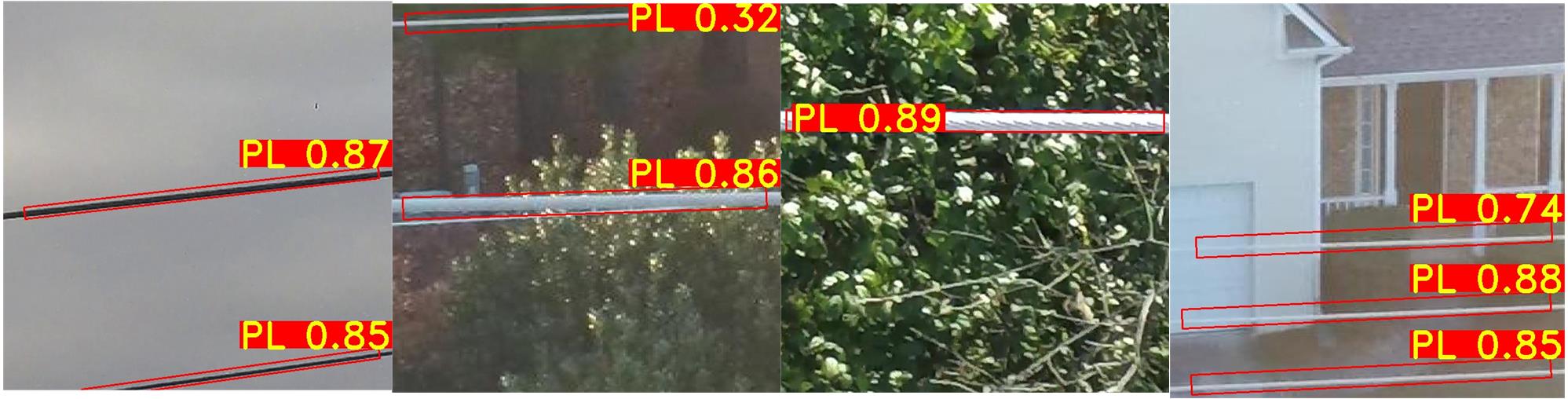}%
\label{fig11_three_case}}
\hfil
\hspace*{-3mm}
\subfloat[]{\includegraphics[width=0.48\linewidth]{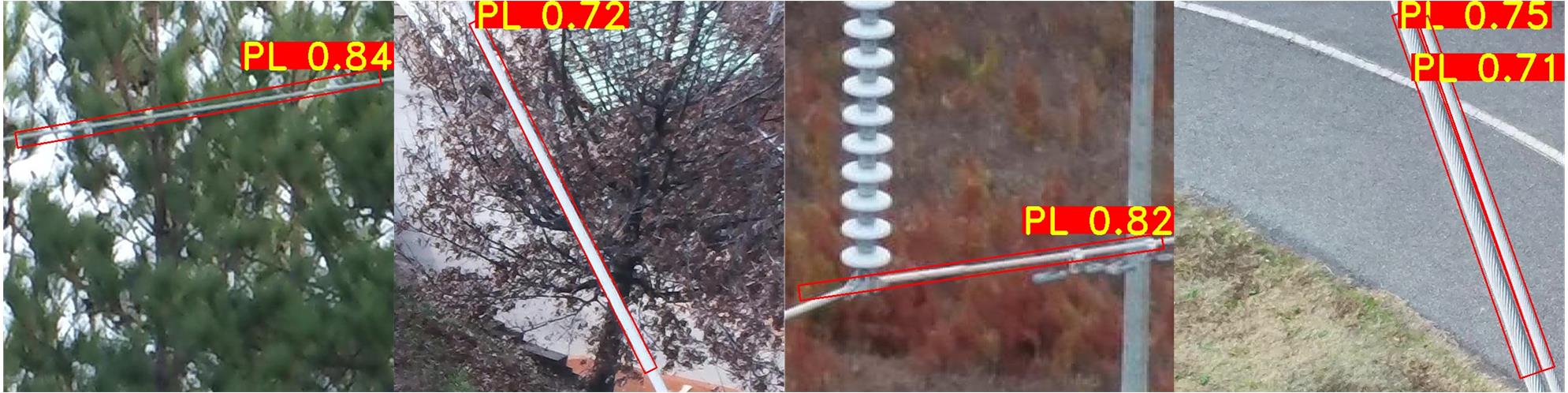}%
\label{fig11_four_case}}
\hfil
\hspace*{3mm}
\subfloat[]{\includegraphics[width=0.48\linewidth]{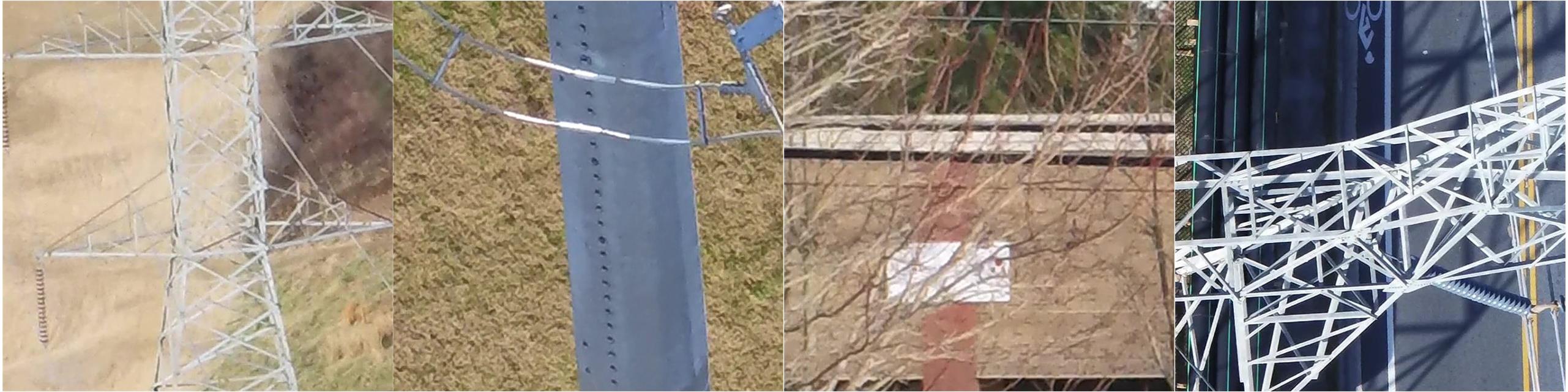}%
\label{fig11_five_case}}
\caption{Examples of detection results using various models and datasets. (a) HBB and mask detection with the Yoloct++(ResNet101) model on TTPLA dataset. (b) OBB detection with YOLOv8(\textit{l}) model on TTPLA-Tile-OBB dataset. (c) OBB detection with PL-YOLOv8(\textit{l}) model on TTPLA-Tile-OBB dataset. (d) Exceptional OBB detection with PL-YOLOv8(\textit{l}) model. \textcolor{blue}{(e) Power lines that are not detected by either YOLOv8 (\textit{l}) or PL-YOLOv8 (\textit{l}) models.}}

\label{fig_11}
\end{figure*}

\textcolor{blue}{The models are compared with the Yolact++ models with backbones of ResNet50 and ResNet101 \cite{ref50}. Yolact++ is an upgraded variation of Yolact, a member of the YOLO family that enhances the YOLO architecture by adding a mask branch to perform instance segmentation tasks.} The training cases and evaluation results are presented in Table \ref{table2}. Their detection results of power lines are showcased in Fig. \ref{fig_11}. The performance metrics used across the models include $AP^{50}$ and $AP^{50-95}$. The $AP^{50}$ represents average precision at an IoU (intersection over union) threshold of $50\%$, while $AP^{50-95}$ computes the average precision at IoU thresholds ranging from $50\%$ to $95\%$, in $5\%$ increments. 

\textcolor{blue}{The Yolact++ model can detect both traditional HBB and masks, providing a baseline for UAV aerial image-based power line detection \cite{ref43,ref50}. While masks allow pixel-level detection of power lines, their accuracy is relatively low, as shown in TABLE \ref{table2}. The quality of masks is not high with discontinuous pixels and missing detections, as shown in Fig. \ref{fig11_first_case}. Comparatively, the HBB tasks have a relatively higher accuracy, achieving an $AP^{50}$ of $42.32\%$ by Yolact++ (ResNet101). However, it fails to frame power lines with constrained bounds.  }
\begin{table}[!t]
\color{blue}
\caption{Comparison of Different Training Mode For Directional Block}
\label{table_new}
\centering
\begin{tabular}{|c|c|c|c|}
\hline
\textbf{Models} & \textbf{Directional block} & \hspace{0.35cm} $\mathbf{AP^{50}}$\hspace{0.35cm} & \hspace{0.12cm} $\mathbf{AP^{50-95}}$ \hspace{0.12cm}\\ \hline
PL-YOLOv8(\textit{s})      & Frozen         & 76.30\%          & 40.83\%             \\ \hline
PL-YOLOv8(\textit{s})     & Trainable       & 78.14\%          & 44.32\%             \\ \hline
PL-YOLOv8(\textit{s})      & Mixed         & 77.26\%          & 43.67\%             \\ \hline
\end{tabular}
\color{black}
\end{table}

OBB tasks were anticipated to resolve the limitations observed in mask and HBB tasks. When employing the modified dataset (TTPLA-Tile-OBB) which comprises tiles to train the model, there is a significant improvement in accuracy across all sizes of YOLOv8 (\textit{n, s, m, l}). The $AP^{50}$ increases to $69.00\%$, $76.33\%$, $70.70\%$, and $73.22\%$, respectively. Fig. \ref{fig11_second_case} shows that the YOLOv8 model, using YOLOv8(\textit{l}) as an example, effectively detects power lines with OBBs. However, a missing detection is evident in the second image of Fig. \ref{fig11_second_case} when compared to Fig. \ref{fig11_three_case}. \textcolor{blue}{The proposed advanced PL-YOLOv8(\textit{n}) model improves \(AP^{50}\) by 9.24\% and \(AP^{50-95}\) by 7.48\%,} while PL-YOLOv8(\textit{s, m, l}) models improve \(AP^{50}\) and \(AP^{50-95}\) by 1\% to 2\%. Fig. \ref{fig11_three_case} illustrates the enhanced detection capabilities of these models, as shown by the results using PL-YOLOv8(\textit{l}). These models generally achieve higher confidence scores compared to the standard YOLOv8 and excel at detecting power lines with weak features. 

\begin{figure}[!t]
\centering
\includegraphics[width=0.95\linewidth]{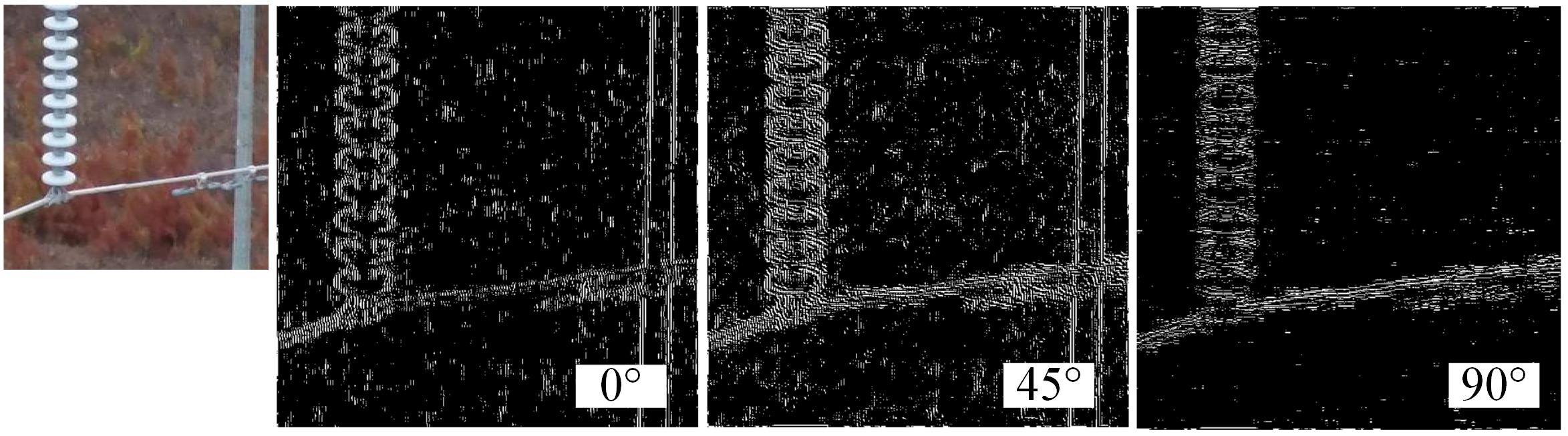}
\caption{Edge enhancement with directional filtering at angles 0°, 45°, 90°. }
\label{fig_12}
\end{figure}

\textcolor{blue}{Additionally, we investigated the impact of directional block enhancement on PL-YOLOv8(\textit{s}) using different training modes. Table \ref{table_new} presents the results. While freezing the directional filter layers ensures consistent directional feature extraction for the deep layers, it results in performance similar to the original YOLO models due to less flexibility. In the mixed training mode, the directional layers are frozen for the initial 1800 epochs and trained for the final 200 epochs. This strategy enables the network to stabilize feature extraction in the early stages, allowing it to later adapt to more complex patterns to enhance model performance. PL-YOLOv8(\textit{s}) in the mixed training mode has a 0.96\% improvement in \(AP^{50}\) and a 2.84\% increase in \(AP^{50-95}\) compared to the frozen mode. Making the directional filter layers fully trainable provides greater adaptability, enabling the initial layers to extract meaningful directional features from the start and resulting in the best performance gains.}  

Fig. \ref{fig11_four_case} showcases additional power lines that were not detected by YOLOv8(\textit{l}). These lines exhibit either non-typical features like those connected to insulators or weak features against a complex or confusing background. The superior performance of the enhanced PL-YOLOv8 models is attributed to directional filters that excel in extracting features from various orientations and complicated textures. Fig. \ref{fig_12} visually displays the binarized textures extracted by these directional filters at $0^\circ$, $45^\circ$, and $90^\circ$, showcasing their effectiveness in feature detection. \textcolor{blue}{Fig. \ref{fig11_five_case} illustrates power line cases that were not detected by either the YOLOv8 (\textit{l}) or PL-YOLOv8 (\textit{l}) models. These cases have weak line characteristics that are either too distant to be detected or are obscured by complex transmission tower structures or tree branches, making detection challenging.} 

\color{blue}
\subsection{Real-time detection capability evaluation}

The real-time detection capability of the proposed models was evaluated by testing their frame rates on an NVIDIA GeForce RTX 4060 Laptop GPU (8188 MiB), as shown in TABLE \ref{table2}. While PL-YOLOv8 models generally produce lower FPS than standard YOLOv8, the PL-YOLOv8(\textit{n}) model maintains a high performance of 97 FPS, which far exceeds the typical real-time threshold of 15–30 FPS. While UAV processors (e.g., Jetson modules) have stricter power and thermal constraints, this performance margin allows effective down-scaling. Further optimizations, such as TensorRT deployment, can enhance FPS performance on embedded platforms (e.g., Nvidia Jetson Xavier NX). This evaluation demonstrates the proposed model’s capability for accurate and real-time power line detection, showing its potential for UAV field applications.
\color{black}

\subsection{Evaluation of vegetation encroachment metric method}

\begin{figure}[!t]
\centering
\includegraphics[width=1\linewidth]{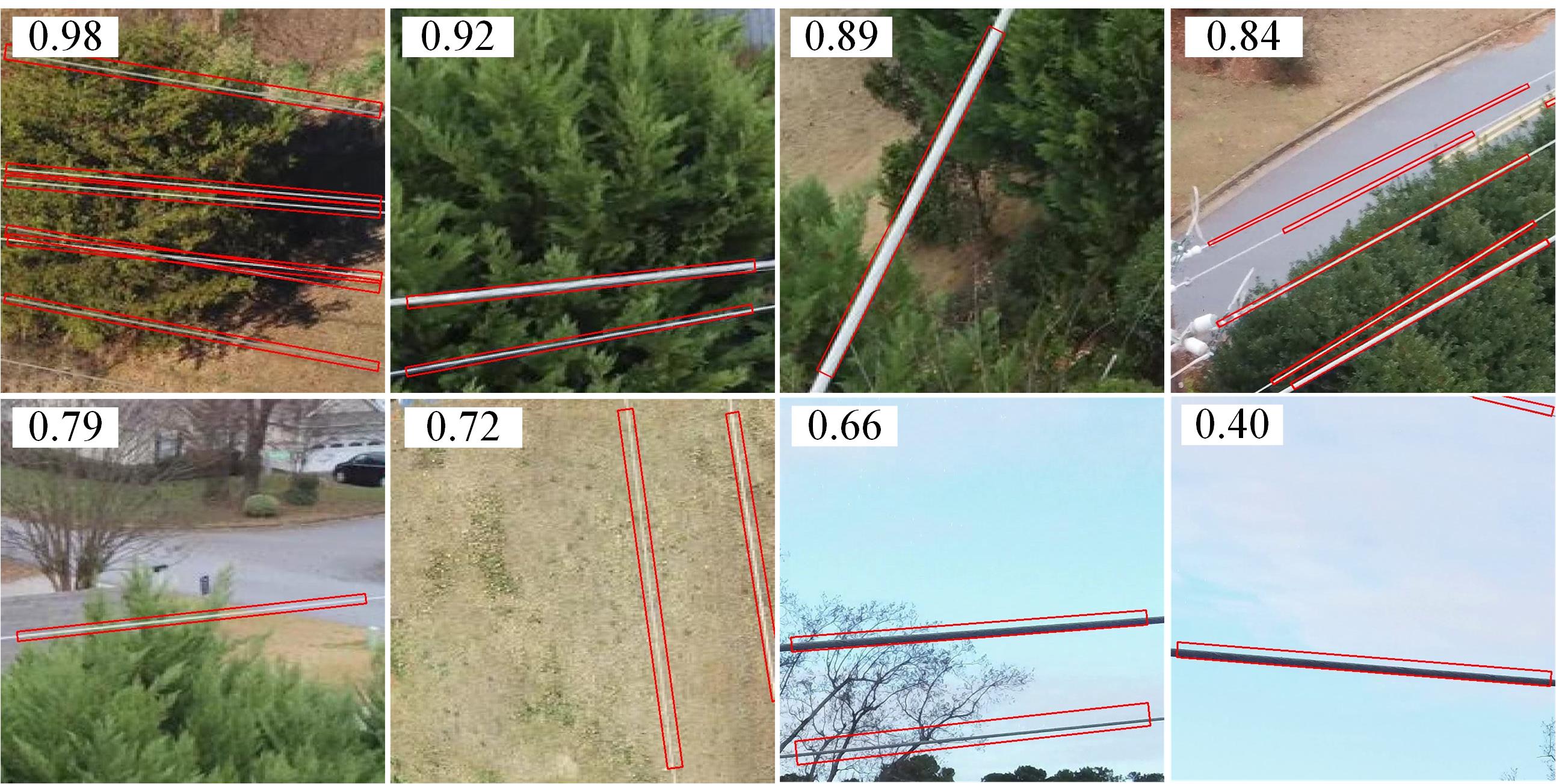}
\caption{Vegetation encroachment metrics: sorted image examples.}
\label{fig_13}
\end{figure}

\begin{figure}[t]
    \centering
    \hspace*{-5mm}
    \subfloat[]{
        \includegraphics[width=0.8\linewidth]{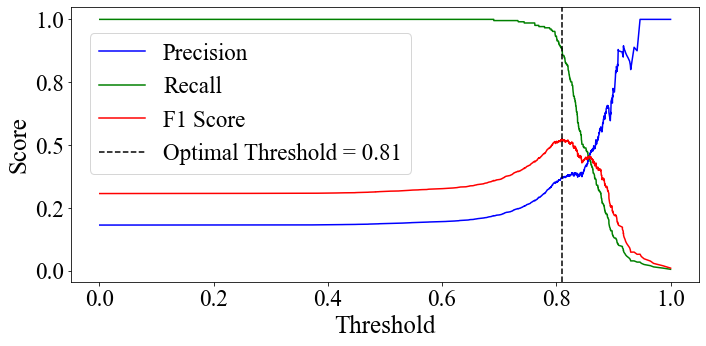}
        \label{fig14_first_case}
    }
    \hfil
    \hspace*{-5mm}
    \subfloat[]{
        \includegraphics[width=0.8\linewidth]{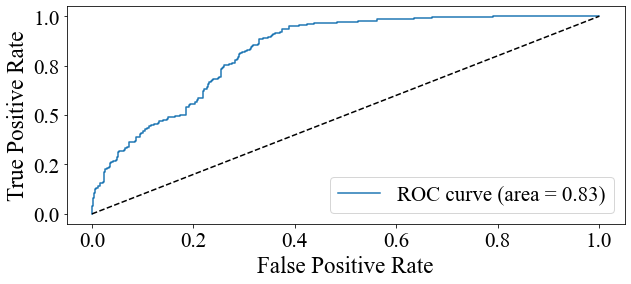}
        \label{fig14_second_case}
    }
    \caption{Evaluation of the vegetation metric method: (a) Precision, recall, and F1 scores over different thresholds. (b) ROC curve and AUC.}
    \label{fig_14}
\end{figure}

\begin{table}[t]
\caption{Evaluation Metrics at Optimal Threshold\label{table4}}
\centering
 \arrayrulecolor{black}
\begin{tabular}{|c||c|c|c|c|}
\hline
\textbf{Optimal Threshold}&\textbf{Accuracy}& \textbf{Precision}& \textbf{Recall} & \textbf{F1 Score}\\
\hline
0.81 & 0.71 & 0.37 & 0.89 & 0.52 \\
\hline
\end{tabular}
\end{table}

The vegetation encroachment metric method was evaluated using 1,000 images from the test dataset. Vegetation encroachment metrics were calculated for each image by evaluating the proximity of vegetation to power lines, using power line detections from PL-YOLOv8. Examples of vegetation encroachment metrics are shown in Fig. \ref{fig_13}, demonstrating how these values can indicate trends in vegetation encroachment. For evaluation, the continuous metric was converted into a binary classification problem using a threshold. If the metric value was greater than or equal to the threshold, the image was classified as alarming for vegetation encroachment; otherwise, it was not. The ground truth for the images was generated by manually labeling them into two classes: alarm and non-alarm. Based on the ground truth and prediction results, the performance was then evaluated using metrics such as accuracy, precision, recall, and F1 score.

Precision is the ratio of true positives to the total positive detections (both true positives and false positives), reflecting the model's accuracy in identifying actual alarms. Recall, or sensitivity, is the ratio of true positives to all actual positives (true positives and false negatives), showing the model's ability to identify all potential hazards. The F1 score reflects the balance between precision and recall, showing the model's effectiveness in accurately identifying trees at risk of vegetation encroachment while maintaining high sensitivity and precision to minimize false alarms.

\begin{figure}[!t]
\color{blue}
\centering
\includegraphics[width=0.8\linewidth]{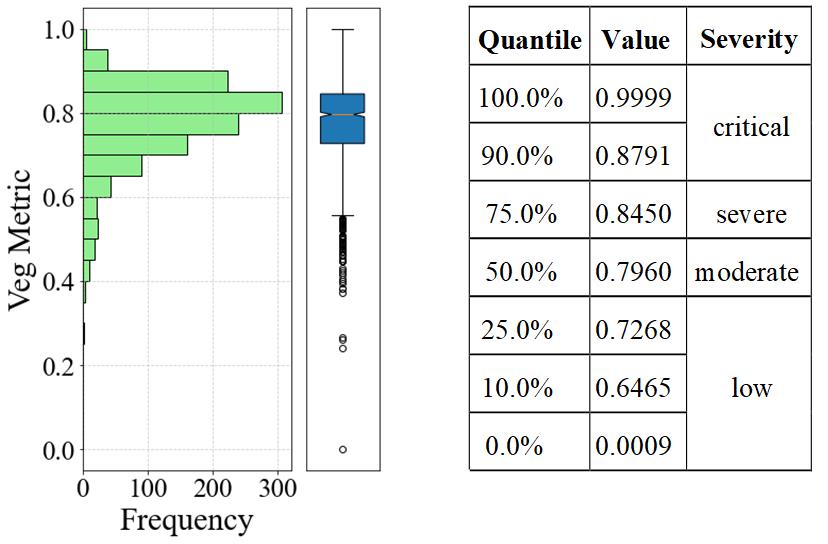}
\caption{Severity levels based on metric percentiles.}
\label{fig_15}
\end{figure}
\color{black}

To find the optimal threshold, we plotted the precision, recall, and F1 scores as shown in Fig. \ref{fig14_first_case}. The optimal threshold was determined to be 0.81, where the F1 score is highest.  The evaluation metrics at this threshold are presented in Table \ref{table4}. \textcolor{blue}{With this threshold, the higher recall is prioritized over precision, aligning with the goal that false negatives are more critical than false positives in vegetation management since false negatives (missed encroachments) could result in severe consequences such as power outages, fires, or safety hazards. This ensures that most encroachment cases are identified, reducing the risk of missing high-priority areas requiring immediate attention.} Fig. \ref{fig14_second_case} illustrates the Receiver Operating Characteristic (ROC) curve, which displays the performance of the classification model across all thresholds. This curve plots the true positive rate against the false positive rate. The Area Under ROC Curve (AUC) indicates the model's ability to distinguish between positive (alarm) and negative (non-alarm) classes across different thresholds. An AUC of 0.83 demonstrates the model's high discrimination capability, effectively distinguishing between images with and without alarms. This result demonstrates the effectiveness of the proposed vegetation encroachment metric in effectively identifying potential risks.

\color{blue}
The evaluation demonstrates that the proposed metric effectively captures the severity of vegetation encroachment. Therefore, in addition to relying on a hard predefined threshold, we also propose a level-based method to indicate severity and prioritize risk areas. Fig. \ref{fig_15} illustrates the distribution of the metric data and provides an example of the severity table, which is based on the metric percentiles:

\begin{itemize}
    \item Critical: greater than the 90th percentile ($>0.8791$).
    \item Severe: between the 75th and 90th percentiles ($0.8450 - 0.8791$).
    \item Moderate: between the 50th and 75th percentiles ($0.7961 - 0.8450$).
    \item Low: below the 50th percentile ($<0.7961$).
\end{itemize}

This approach avoids using a hard threshold and is particularly useful for datasets lacking ground truth for vegetation encroachment. It can provide utilities with severity levels to prioritize maintenance schedules, optimize resource allocation, and identify high-risk areas for proactive intervention.  
\color{black}

\section{Conclusion}

We proposed an advanced YOLO-based deep learning framework for effectively detecting power lines and evaluating vegetation encroachment using UAV aerial images. The framework offers a PL-YOLOv8 neural network, which enhances the traditional YOLOv8 with an innovative directional block for efficiently extracting weak features of thin and long power lines and their vicinity in complex backgrounds. Additionally, it is the first to use OBBs to detect power lines, providing precise localization with high accuracy. The comparative analysis indicates that the proposed PL-YOLOv8(\textit{n}) model achieves an \(AP^{50}\) of $78.24\%$, and a frame rate of 97 FPS.

In addition, a vegetation encroachment metric method is introduced to assess the proximity of nearby vegetation to power lines based on the OBB location. This approach offers a straightforward and effective alternative to complex segmentation tasks for tree detection by leveraging image RGB indices, textures, and brightness to assess vegetation encroachment. The results demonstrate promising effectiveness, with an accuracy of 0.71 and an F1 score of 0.52 in our dataset.

\vspace{-10pt}
\begin{IEEEbiographynophoto}
{Shuaiang Rong} (Student Member, IEEE) received the M.S. degree in electric engineering from the Shanghai University of Electric Power, China, in 2018. She is currently working toward the Ph.D. degree at the University of Illinois, Chicago. Her research interests include deep learning, image processing, renewable energy integration, and power system protection.
\end{IEEEbiographynophoto}
\vspace{-10pt}
\begin{IEEEbiographynophoto}%
{Lina He} (IEEE Senior Member) received her Ph. D. degree in electrical engineering at University College Dublin, Ireland in 2014. She is currently an Assistant Professor in department of electrical and computer engineering, University of Illinois at Chicago. She was a Project Manager and Senior Consultant at Siemens headquarter in Germany and Siemens US from 2014 to 2017. Her research interests include power system resilience, renewable energy integration, power system security and operation, and AI applications in power systems.
\end{IEEEbiographynophoto}
\vspace{-10pt}
\begin{IEEEbiographynophoto}%
{Salih Furkan Atici} received the Ph.D. and M.S. degrees in electrical and computer engineering from the University of Illinois Chicago, Chicago, USA, in 2020 and in 2024, respectively. He is currently working at Apple Inc. in Cupertino, California. His research interests include deep learning, image processing, video processing, video codec, and machine learning.
\end{IEEEbiographynophoto}
\vspace{-10pt}
\begin{IEEEbiographynophoto}%
{A.E. Cetin} (Fellow, IEEE) received his B.Sc. from METU, Ankara, Turkey, and Ph.D. in 1987 from the University of Pennsylvania, USA. He was an Assistant Professor at the University of Toronto between 1987-1989 in Canada. He was a faculty member at Bilkent University from 1987-2017. He is currently a professor in the Department of Electrical and Computer Engineering at the University of Illinois at Chicago (UIC). He also has held visiting professor positions at Bellcore (1988), University of Minnesota (1996-1997), and UC San Diego (2016-2017). He has been carrying out research in the areas of theoretical and applied machine learning, signal, image, and video processing, biomedical signal processing, infrared and chemical sensor signal processing in Cyber-Physical Systems (CPS). His group introduced the concept of adaptive prediction and split vector quantization for Line Spectral Frequency representation. This concept was used in ITU speech coding standards including G.729, G.723.1 and GSM EFR.

He became Fellow of IEEE for his contributions to signal and image recovery in 2010. He is the Editor-in-Chief of Signal, Image and Video Processing, Springer-Nature. He received a best paper award for his camera-based wildfire detection work in a conference organized by UNESCO and Cyprus Presidency of the European Union.  He is one of the co-founders of the multinational smart wide angle OEM camera company Oncam-Grandeye, UK. He served as the CEO/CTO of Grandeye, Turkey between 2003-2013. Oncam-Grandeye cameras won design and innovation awards in IFSEC, UK, and ISC WEST, Las Vegas, trade fairs.
\end{IEEEbiographynophoto}
\vspace{-10pt}

\vfill

\end{document}